\documentclass{egpubl}
\usepackage{eg2026}

\ShortPresentation      
\usepackage[T1]{fontenc}
\usepackage{dfadobe}  
\usepackage{makecell}
\usepackage{multirow}
\usepackage{float}
\usepackage{xcolor}
\usepackage{colortbl}

\usepackage{cite}  
\BibtexOrBiblatex
\electronicVersion
\PrintedOrElectronic
\ifpdf \usepackage[pdftex]{graphicx} \pdfcompresslevel=9
\else \usepackage[dvips]{graphicx} \fi

\usepackage{egweblnk} 

\title[VkSplat]%
      {VkSplat: High-Performance 3DGS Training in Vulkan Compute}

\author[J. Chen, M. Ibrahim, \& Y. Liu]
{
\parbox{\textwidth}{\centering
    Jingxiang Chen$^{1}$ \orcid{0009-0007-1591-2909}
    , Mohamed Ibrahim$^{1}$ \orcid{0000-0002-3559-3761}
    , and Yang Liu$^{1}$ \orcid{0009-0003-7469-8608}
\thanks{Chairman Eurographics Publications Board}
        }
        \\
        \centering $^1$Huawei Canada
       }

\begin{document}


\maketitle
\begin{abstract}
   We present VkSplat, a high-performance, cross-vendor 3D Gaussian Splatting (3DGS) training pipeline implemented fully in Vulkan compute, addressing performance and compatibility limitation of existing training pipelines. With various optimizations, we achieve $3.3\times$ speed and $33\%$ VRAM reduction over CUDA+PyTorch baseline, maintaining quality, and demonstrating compatibility across GPU vendors. To the best of our knowledge, this is the first fully-Vulkan-based 3DGS training pipeline that achieves state-of-the-art performance. Code: \href{https://github.com/harry7557558/vksplat}{https://github.com/harry7557558/vksplat}
   \\


\end{abstract}  
\section{Introduction}

Since its introduction \cite{kerbl3Dgaussians}, 3D Gaussian Splatting (3DGS) has been widely adopted for fast high-quality novel view synthesis. Despite its wide application, training time and memory usage remain barriers for practical use, especially for large scenes and applications requiring timely training. While novel algorithms have been introduced to improve 3DGS quality and capability over wide range of applications, focus on efficiency is often secondary. Existing 3DGS training implementations \cite{kerbl3Dgaussians} \cite{ye2025gsplat} are often sub-optimal in performance, as well as CUDA ecosystem blocking deployment across hardware vendors.

To address these limitations, we present VkSplat, a high-performance, cross-vendor 3DGS training pipeline implemented fully in Vulkan compute. We introduce complete tile culling using a parallelizable scan-line formulation, per-Gaussian and tensor-based rasterization backward eliminating pixel-level atomic contention, and fully-fused backward projection and Adam optimizer in a single pass. Our end-to-end Vulkan implementation outperforms state-of-the-art CUDA pipelines, achieving $3.3\times$ speed up over GSplat \cite{ye2025gsplat}, 33\% VRAM reduction, and identical image quality, as well as being compatible across GPU vendors.

\section{Related works}

3D Gaussian Splatting (3DGS) was introduced by \cite{kerbl3Dgaussians}, which involves representing a 3D scene using a set of Gaussian ellipsoids, typically trained through differentiable rendering using gradient descent. Each Gaussian in world-space is parameterized by mean $\mu$ interpreted as position, a covariance matrix $\Sigma$ commonly parameterized by per-axis log scales and rotation quaternion, an opacity value typically in logit space, and color parameters as spherical harmonics (SH) coefficients typically with degree 3. Most of the existing works \cite{kerbl3Dgaussians} \cite{ye2025gsplat} implement 3DGS training in CUDA and run on NVIDIA GPUs, typically using a tile-based rasterization pipeline consisting of projection, tile binning and sorting, rasterization, backward passes for rasterization and projection, optimizer, and often densification stages.

Training a 3DGS scene can take tens of minutes to hours depending on scene complexity and number of Gaussians. Training speed can be improved by reducing the number of false-positive intersections and therefore accelerate sorting and rasterization, like introduced in \cite{radl2024stopthepop} \cite{HansonSpeedy} \cite{liao2025litegs}. The rasterization backward step is often a bottleneck of 3DGS rasterization, and optimized implementations have been proposed by \cite{taming3dgs} \cite{liao2025litegs}. Adam optimizer is also a large performance bottleneck with large number of Gaussians, which is addressed by \cite{taming3dgs}.

In addition to training time, reliance on CUDA and PyTorch in existing implementations have limited 3DGS training to NVIDIA GPUs. Attempts have been made to use cross-vendor API like Vulkan for rasterization: \cite{park24vkgs} \cite{yuan2025efficient} use the graphics pipeline for rasterization, but no end-to-end training was done. \cite{müller2025momentbased3dgs} claims to have a Slang+Vulkan differentiable rendering pipeline, but PyTorch is used for training. To the best of our knowledge, our work is the first end‑to‑end 3DGS training pipeline implemented entirely in Vulkan, free of any NVIDIA‑specific extensions or dependencies, while delivering performance that surpasses CUDA‑based baselines by a substantial margin.

\section{Pipeline overview}

The design of VkSplat focuses on high performance, cross-vendor training, and fidelity consistent with academic baselines. For the GPGPU API, we choose Vulkan, a modern framework that offers high performance and support for mainstream GPU vendors.
We based our optimizations on Slang-Gaussian-Rasterization \cite{kopanas24slanggs}, a differentiable 3DGS renderer based on Slang shading language. A strength of Slang is its ability to target multiple backends. Although the original implementation was designed for CUDA and PyTorch, we successfully run it with a Vulkan backend and preserve the flexibility to support additional backends in the future.

\section{Key technical contributions}

In addition to introducing a cross-vendor, minimal-dependency Vulkan-based 3DGS training pipeline, numerous optimizations are made to push performance exceeding CUDA baselines while maintaining fidelity. Notable contributions are summarized below.

\subsection{Complete tile culling using scan-line intersection}

In tile-based rendering, a list of Gaussian-tile intersections is generated and traversed.
Existing implementations produce false-positive intersections:
\cite{kerbl3Dgaussians} determines intersections based on distance between tile and Gaussian center with a conservative radius; \cite{radl2024stopthepop} uses tiles overlapping with tight bounding boxes of Gaussians, which still produces false positives; \cite{ren2025fastgs} introduces a Compact Box approach, which produces false negatives. False-positive intersections drop performance in sorting and rasterization, and false-negative intersections lead to rendering inaccuracy. \cite{radl2024stopthepop} culls false positive intersections in a separate pass, but this introduces computation time and VRAM usage overhead.

\begin{figure}
    \centering
    \includegraphics[width=\linewidth]{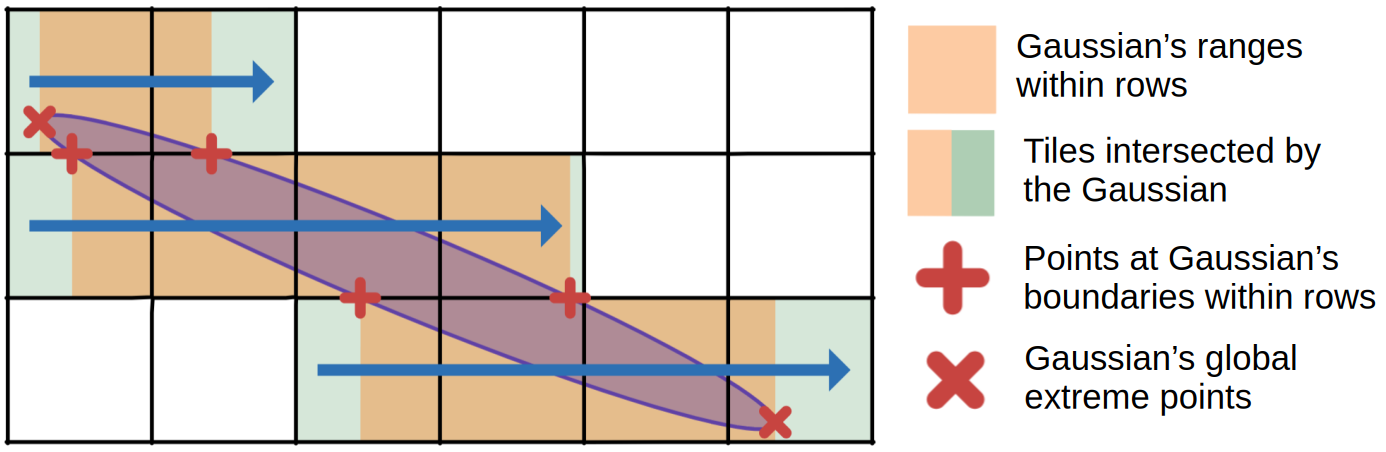}
    \caption{Given a screen-space Gaussian ellipse, we first pick the shorter dimension (vertical in the case). For each row (or column) of tiles, we first find the ellipse's range of coordinates within the row, which is bounded by either the boundary or global extreme points of the ellipse. We implement the closed-form solution as a highly optimized branchless function. By rounding down the minimum coordinate and rounding up the maximum coordinate toward tile boundary, we are able to efficiently count or iterate through tiles overlapped by the ellipse without false positive or negative.}
    \label{fig:tile-culling}
    \vspace{-15pt}
\end{figure}

In our implementation, we compute number of the intersecting tiles per Gaussian in the projection-forward pass, and use a separate pass to fill the buffer with tile-depth pairs. Inspired by \cite{HansonSpeedy} \cite{liao2025litegs}, we compute exact intersections using a scan-line approach.
Given a Gaussian represented as an ellipse containing opacity above threshold and an interval for one axis, we compute closed-form interval on other axis that intersect the ellipse, allowing efficient intersection counting and traversal, as shown in Figure \ref{fig:tile-culling}.

\subsection{Rasterization backward with adaptive scheduling}

In existing implementations \cite{kerbl3Dgaussians} \cite{ye2025gsplat}, rasterization backward launches threads per pixel, which atomically adds gradient to each Gaussian from each pixel, resulting in high contention from atomics and subgroup reduction. To address this, we switch between two rasterization backward implementations.

Our first implementation was inspired by \cite{taming3dgs}, which parallelizes threads over Gaussians instead of pixels to reduce atomic contention. In our implementation, instead of using Gaussian batches fixed to warp size, we launch one thread block per tile and dynamically adjust Gaussian batch size based on number of Gaussians binned to the tile. For a tile with $P$ pixels and $N$ Gaussians, with Gaussian batch size $S$, the latency is approximately proportional to the product of number of batches $(P+S)/P$ and number of Gaussians traversed per batch $(N+S)/S$, minimized by $S=\sqrt{NP}$. We choose $S$ by rounding up $\sqrt{NP}$ to a multiple of subgroup size, capped to $128$ that balances performance and hardware occupancy, which empirically produces the lowest latency.

Our second implementation aims to improve thread utilization and minimize divergence. For a Gaussian batch, we first run a forward pass parallelized over pixels, compute transmittance and its derivative with respect to Gaussian parameters for each Gaussian-pixel pair, and store them in shared memory. Then, we run a backward pass paralellized over Gaussians, fetch pre-computed transmittance and derivatives from shared memory, and compute accumulated gradient. We parameterize projected Gaussians following \cite{liao2025tcgs}, which simplifies opacity computation into matrix multiplication and allows efficient parallel computation.

In our benchmark, we notice the fastest implementation depends on training configuration. For example, on NVIDIA RTX 3090, the second implementation is 20\%-30\% faster than the first on Mip-NeRF 360 bicycle scene, but the first one is faster on garden scene. To automatically find the fastest implementation, we use a Thompson sampling scheduler, which stores a latency belief of each implementation as distributions, randomly selects an implementation with higher probabilities for faster ones, and updates belief based on measured latency.
This improves performance over using a fixed implementation when a faster implementation is available, with minimal overhead when the fixed implementation is already faster.

\begin{figure}
    \centering
    \includegraphics[width=\linewidth]{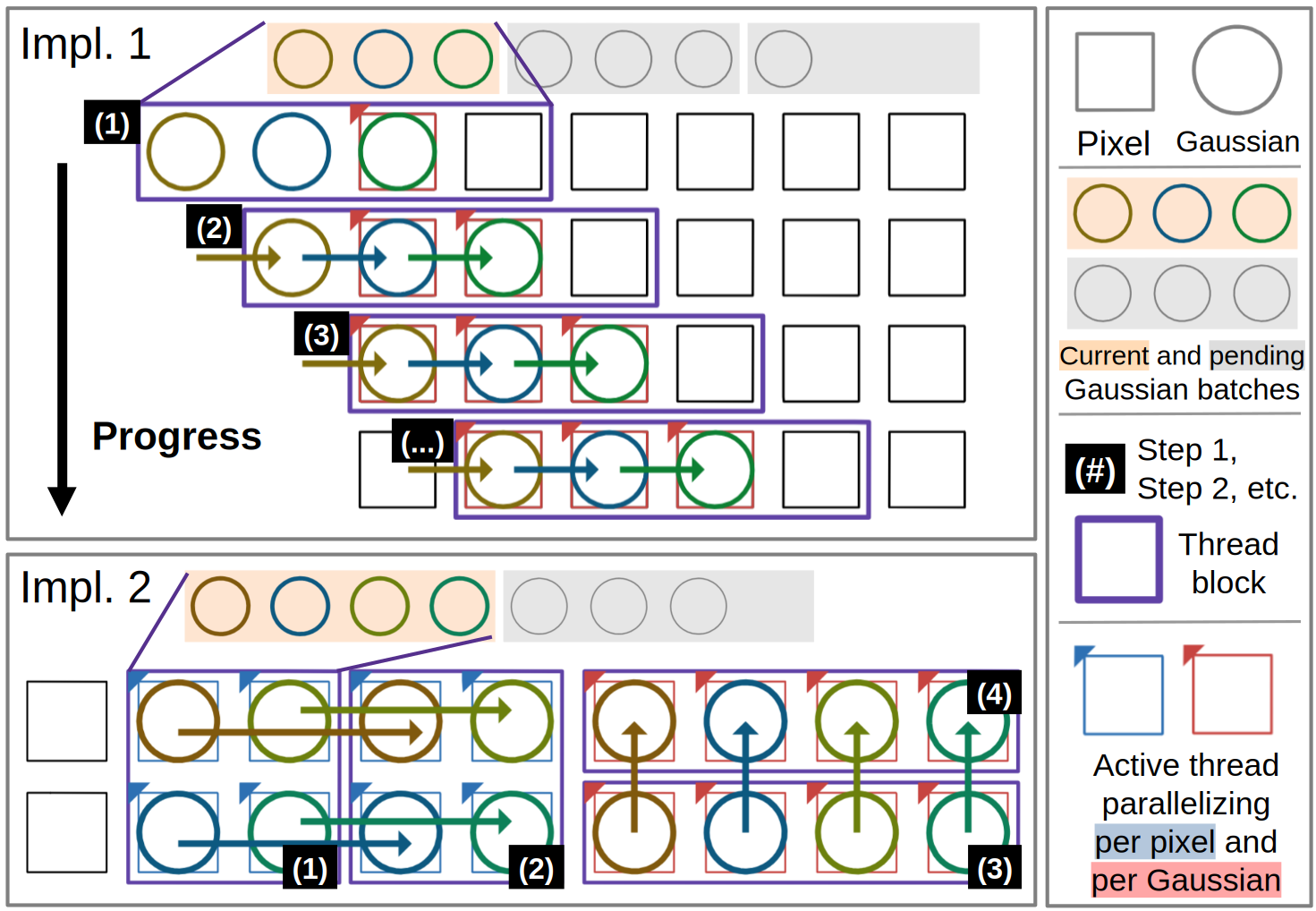}
    \caption{Two rasterization backward implementations. The first implementation is similar to \cite{taming3dgs}, except using one thread block per tile in a single pass, and dynamically adjusting Gaussian batch size based on number of Gaussians binned to the tile. The second implementation first performs a forward pass with per-pixel parallelization and stores transmittance and its derivative in shared memory, then performs gradient accumulation with per-Gaussian parallelization, utilizing all threads with minimum divergence.}
    \label{fig:placeholder}
    \vspace{-15pt}
\end{figure}

\subsection{Fused projection backward and optimizer}

Most existing 3DGS training implementations use Adam optimizer from PyTorch, which is not fused by default and incur unnecessary memory footprint. The largest Gaussian attribute, SH coefficients, involves separate learning rates for degree 0 and the rest of parameters, and concatenating large tensors in existing implementations \cite{kerbl3Dgaussians} \cite{ye2025gsplat} unnecessarily increase memory footprint. \cite{taming3dgs} addresses this issue by avoiding concatenation, but the gradients of Gaussian parameters are still stored. Storing log and logit mapping of scales and opacities also introduces memory footprint.
In VkSplat, we fuse projection backward and Adam optimizer into a single kernel and transform value and gradient of scales and opacities to log and logit space within the optimizer, which completely eliminates aforementioned memory footprint.

\subsection{32-bit tile-depth pair sorting}

Sorting implementations in Vulkan \cite{park24vkradixsort} are often slower than CUDA's heavily engineered built-in sorting. To address this limitation, we use 32-bit sorting keys that involve tile ID in higher bits and depth in lower bits. We map depth using $z \rightarrow (2z + 1) / (z + 1)$, which transforms positive real values into values between 1 and 2 (exclusive), in a way that only the lower 23 mantissa bits are different when represented in FP32. Instead of following popular CUDA implementations that use 64-bit sorting keys with tile ID in higher 32 bits and depth in lower 32 bits, we use a 32-bit sorting key with tile ID in minimum number of higher bits, and place higher bits of floating point mantissa in lower bits. On images with resolution up to 1080p with $16\times16$ tile size, this leads to at most 14 bits for tile ID and at least 18 bits for depth. We did not notice any difference in quality metrics compared to using 64-bit sorting.

\subsection{Fully-fused loss gradient evaluation}

3DGS training conventionally use a weighted sum of $L_1$ and SSIM losses. Existing 3DGS implementations \cite{kerbl3Dgaussians} \cite{ye2025gsplat} explicitly evaluate losses in PyTorch and use fused-ssim \cite{taming3dgs} for SSIM loss, which operates on channel-first memory layout that requires conversion from channel-last 3DGS renders. In our implementation,
we use a fully-fused kernel that directly computes the gradient of weighted sum of $L_1$ and SSIM losses in a single pass, without intermediate reduction or memory layout conversion. Also, we store the reference image in $4\times$UINT8 RGBA pixels, without conversion into FP32 that introduce additional memory overhead. Support for alpha masking is also fused with near-zero overhead.

\subsection{Additional features}

We implement initialization and two densification strategies, default \cite{kerbl3Dgaussians} and MCMC \cite{kheradmand20243d}, closely following \cite{ye2025gsplat} but in as few shader launches as possible. Degree 3 SH include 48 FP32 coefficients per Gaussian, which we split into 12 128-bit values in a column-based format aligned with subgroup size, offering improved memory coalescing. Likewise, we also fuse scale and opacity into a single 128-bit value.

\section{Results}

We evaluate VkSplat on 7 permissively released scenes from the Mip-NeRF 360 dataset \cite{barron2022mipnerf360} and report mean quantitative results across datasets. Following \cite{ye2025gsplat}, we downscale images with the highest resolution and save in lossless 8-bit PNG format. We use image resolution, validation image selection, and training hyperparameters consistent with default of \cite{ye2025gsplat}.

\subsection{Quality}

Since 3DGS training is stochastic, we trained each scene with each method 5 times and reported metrics (PSNR, SSIM, LPIPS evaluated consistent with \cite{ye2025gsplat}, and number of Gaussians) with 90\% confidence interval. Our implementation produces quality consistent with baseline, as shown in Table \ref{tab:quality}.

\begin{table}
    \centering
    \caption{Metrics of GSplat \cite{ye2025gsplat} and our VkSplat with uncertainty. Each implements default \cite{kerbl3Dgaussians} and MCMC \cite{kheradmand20243d} densification strategies. Number of Gaussians is in millions.}
    \label{tab:quality}
    \begin{tabular}{|c|c|c|c|c|}
        \hline
        \thead{Metric} & \thead{GSplat\\Default} & \thead{VkSplat\\Default} & \thead{GSplat\\MCMC} & \thead{VkSplat\\MCMC} \\
        \hline
        PSNR & 29.[19-25] & 29.2[0-7] & 29.4[3-5] & 29.[39-45] \\
        SSIM & 0.87[8-9] & 0.87[8-9] & 0.881 & 0.881 \\
        LPIPS & 0.124 & 0.12[4-5] & 0.1[29-30] & 0.130 \\
        \makecell{NumGS} & 3.0[6-8] & 3.0[2-6] & 1.00 & 1.00 \\
        \hline
    \end{tabular}
\end{table}

\subsection{Resource usage}

We benchmarked our implementation against the baseline \cite{ye2025gsplat}  (commit \texttt{b60e917}) on an NVIDIA RTX 3090 GPU. We queried each stage of the 3DGS training and reported corresponding timing, averaged across the 7 scenes, presented in Table \ref{tab:resource}. As can be seen, we are over $3.3\times$ faster than the baseline, using around 33\% less VRAM. We are also faster than the baseline in every single step of the pipeline. Our complete tile culling accelerates rasterization, added on top of our optimized rasterization backward. Fused projection backward and optimizer and efficient memory layout of SH coefficients speed up projection forward, projection backward, and optimizer by multiple times. With fused kernels, loss gradient evaluation and densification is also multiple times faster. GSplat has large time unaccounted by queries; profiling shows this is largely backward of SH tensor concatenation and small kernel launches managed by PyTorch.

\begin{table}
    \centering
    \caption{Resource usage of VkSplat and baseline, including overall time/VRAM and timing breakdown of each step in seconds. In VkSplat, projection backward and optimizer are fused into one step.}
    \label{tab:resource}
    \begin{tabular}{|c|c|c|c|c|}
        \hline
        \thead{Metric} & \thead{GSplat\\Default} & \thead{VkSplat\\Default} & \thead{GSplat\\MCMC} & \thead{VkSplat\\MCMC} \\
        \hline
        Total Time [s] & 1384 & 412 & 995 & 285 \\
        Total VRAM [GiB] & 4.56 & 3.01 & 1.37 & 0.93 \\
        \hline
        Projection Fwd & 94 & 19 & 40 & 8 \\
        Tiling/Sorting & 42 & 25 & 41 & 27 \\
        Rasterization Fwd & 69 & 25 & 71 & 25 \\
        Loss & 103 & 32  & 110 & 32 \\
        Rasterization Bwd & 246 & 130 & 268 & 120 \\
        Proj Bwd + Optim & 61+398 & 131 & 33+172 & 46 \\
        Densify & 31 & 5 & 61 & 3 \\
        Unaccounted & 341 & 43 & 200 & 23 \\
        \hline
    \end{tabular}
\end{table}

\subsection{Cross-vendor compatibility}

To demonstrate our capability to perform on GPUs without CUDA support, we trained bicycle scene with default densification on both NVIDIA RTX 3090 and AMD Radeon RX 7800 XT, on respectively Windows 11 and Ubuntu 24.04. For both GPUs, we produced consistent quality metrics, VRAM usage, and number of Gaussians. Training took 575 seconds on RTX 3090 and 1201 seconds on Radeon RX 7800 XT. We notice the step with the largest speed differences is loss computation (24s vs 303s) since transferring reference image from host to device appears to be magnitudes slower on AMD. Memory-bound tasks like projection backward/optimizer and projection forward are over twice as slow, but the gap is less for compute-bound tasks like rasterization backward. We believe the gap can be reduced with hardware-specific optimizations.

\section{Discussion and limitations}

While we mainly tested with Vulkan, due to the versability of Slang, this work can be extended to other backends like CUDA, Metal, DirectX, WebGPU, etc., offering compatibility across hardware and platforms. Since the same 3DGS pipeline components (tile culling, rasterization, optimizer) are shared with other splatting variants, our optimizations can be easily extended to other splatting pipelines and benefit wide range of applications.
However, our implementation current lacks practical features for real-world datasets such as exposure correction, depth/normal supervision, batching, multi-GPU training, etc. that are implemented in mature 3DGS trainers \cite{kerbl3Dgaussians} \cite{ye2025gsplat}. Also, we closely followed \cite{ye2025gsplat} in densification \cite{kerbl3Dgaussians} \cite{kheradmand20243d} and currently do not support more efficient densification strategies \cite{taming3dgs} \cite{liao2025litegs} 
. Additional features and more efficient densification can be implemented into VkSplat with engineering effort.

\section{Conclusion}

We presented VkSplat, an end-to-end 3DGS training pipeline implemented fully in Vulkan compute, which demonstrates careful GPU optimization yielding substantial performance gains, achieving $3.3\times$ speedup, using 33\% less VRAM, being cross-platform, while remaining high-fidelity.
We publicly release our code for reproducibility and community benefit.

\bibliographystyle{eg-alpha-doi} 
\bibliography{egbibsample}       

\end{document}


\maketitle

\section{Reproduced metrics}

We evaluated VkSplat on 7 permissively released scenes from Mip-NeRF 360 dataset, using images downscaled 4x and 2x for respectively outdoor and indoor scenes and saved in lossless PNG format, consistent with GSplat. Each setup is benchmarked 5 times on a NVIDIA RTX 3090 GPU, and 90\% confidence interval of mean is reported, rounded consistent with existing academic reports.

For quality metrics, we report PSNR, SSIM, and LPIPS evaluated using two methods. LPIPS Alex is evaluated consistent with GSplat, which we use as our baseline in the paper. LPIPS VGG is evaluated consistent with original Inria implementation, which is more widely used in existing academic reports. 

We report two VRAM values: total VRAM that intends to be fair when comparing with VRAM reported by GSplat, which is used by the paper but is lower than the actual VRAM reported by \texttt{nvidia-smi}; and peak VRAM that more accurately reflects whether an OOM (Out of Memory) will happen. The total VRAM is computed by summing all allocated buffers, while peak VRAM accounts for VRAM spike during buffer resizing. The choice of having two measures is to provide one value for fair comparison with GSplat (and other popular academic works) that reports VRAM returned by \texttt{torch.cuda.max\_memory\_allocated()}, which does not account for VRAM reserved by PyTorch caching allocator and allocated from custom CUDA code; and another value that more accurately reflects practical usability.

To ensure fairness of VRAM reporting in the paper, we trained bicycle scene with default densification on NVIDIA RTX 3090 GPU with both GSplat and VkSplat, while polling \texttt{nvidia-smi} at 1 Hz to measure maximum VRAM used by the process. For VkSplat, we measured 5.73 GiB maximum VRAM, slightly above reported total VRAM (5.72 GiB) but lower than reported peak VRAM (6.75 GiB). We believe this is because there's small amount of VRAM reserved by Vulkan runtime, but buffer resizing has short duration and was not captured by a sample rate of 1 Hz. Therefore, we assume an actual VRAM usage slightly over $(6.75 / 5.72 - 1) \times100\% = 18\%$ higher than reported in the paper. For GSplat, we measured 11.98 GiB maximum VRAM, while the software itself reports 8.68 GiB VRAM usage, indicating an actual VRAM usage $38\%$ higher than reported in the paper, plus additional VRAM usage not captured by a sample rate of 1 Hz. Since $38\%$ is higher than $18\%$, we do not believe VRAM reporting in the paper introduces an unfair advantage for VkSplat.

In the following tables, time is in seconds, VRAM usage is in GiB, and number of Gaussians is in thousands.

Metrics for default densification:

\begin{tabular}{|l|c|c|c|c|c|c|c|c|}
\hline
 & bicycle & garden & stump & bonsai & counter & kitchen & room & Average \\
\hline
PSNR  & 25.[48-54] & 27.[65-79]  & 26.[87-91] & 32.[17-26] & 29.1[3-5] & 31.[46-63] & 31.[48-74] & 29.2[0-7]  \\
SSIM  & 0.77[2-3]  & 0.87[0-2]   & 0.78[1-3]  & 0.94[7-8]  & 0.916     & 0.933      & 0.927      & 0.87[8-9]  \\
LPIPS Alex & 0.16[1-2]  & 0.07[1-2]   & 0.14[5-6]  & 0.11[4-5]  & 0.14[1-2] & 0.085      & 0.15[2-3]  & 0.12[4-5]  \\
LPIPS VGG & 0.20[2-4]  & 0.10[3-4]   & 0.20[5-7]  & 0.17[5-6]  & 0.17[8-9] & 0.114      & 0.19[4-5]  & 0.168  \\
Time  & 5[76-83]   & 54[4-6]     & 4[55-69]   & 27[3-6]    & 30[1-5]   & 42[0-3]    & 3[05-10]   & 41[2-4]    \\
Total VRAM  & 5.7[0-3]   & 4.9[0-3]   & 4.[58-75]  & 1.3[3-5]   & 1.2[5-6]  & 1.9[2-4]   & 1.[59-64]  & 3.0[5-7]  \\
Peak VRAM  & 6.7[3-6]   & 5.[79-82]   & 5.[40-61]  & 1.5[6-9]   & 1.4[6-8]  & 2.2[6-8]   & 1.[86-92]  & 3.[59-62]  \\
NumGS & 58[21-52]  & [4999-5028] & 4[664-839] & 12[56-78]  & 11[61-88] & 18[52-66]  & 15[52-95]  & 30[55-81]  \\
\hline
\end{tabular}

\newpage

Metrics for MCMC 1M Densification:

\begin{tabular}{|l|c|c|c|c|c|c|c|c|}
\hline
 & bicycle & garden & stump & bonsai & counter & kitchen & room & Average \\
\hline
PSNR  & 25.5[3-8]  & 27.[25-32]  & 26.[89-94] & 32.[59-71] & 29.4[4-9] & 31.[54-79] & 32.[32-46] & 29.[39-45] \\
SSIM  & 0.77[3-4]  & 0.85[4-5]   & 0.7[88-90] & 0.953      & 0.92[3-4] & 0.93[5-6]  & 0.938      & 0.881      \\
LPIPS Alex & 0.184      & 0.105       & 0.16[3-4]  & 0.10[6-7]  & 0.13[0-1] & 0.08[6-7]  & 0.13[2-3]  & 0.130      \\
LPIPS VGG & 0.21[6-7]      & 0.13[4-5]       & 0.21[4-5]  & 0.16[4-5]  & 0.16[4-5] & 0.114  & 0.17[2-4]  & 0.169      \\
Time  & 21[7-9]    & 21[4-6]     & 21[4-5]    & 32[5-7]    & 3[67-70]  & 33[4-6]    & 31[5-6]    & 28[4-5]    \\
Total VRAM  & 0.88       & 0.88        & 0.88       & 0.94       & 0.97      & 0.95       & 0.9[6-7]   & 0.9[2-3]   \\
Peak VRAM  & 0.88       & 0.88        & 0.88       & 0.94       & 0.97      & 0.95       & 0.9[6-7]   & 0.9[2-3]   \\
NumGS & 1000       & 1000        & 1000       & 1000       & 1000      & 1000       & 1000       & 1000       \\
\hline
\end{tabular}

For MCMC densification, we pre-allocate buffers to fit maximum number of Gaussians to avoid buffer resizing, and produced peak VRAM near-identical to total VRAM.

\vspace{1cm}

\section{Timing breakdown}

We benchmarked each setup once and reported timing of each stage. In the paper, tiling/sorting includes index offset, generate keys, sorting, and tile ranges, and loss includes copy image to device and loss gradient. All numbers are in seconds.

Timing breakdown on NVIDIA RTX 3090, Default densification:

\begin{tabular}{|l|c|c|c|c|c|c|c|c|}
\hline
 & bicycle & garden & stump & bonsai & counter & kitchen & room & Average \\
\hline
Projection Forward      & 32.6  & 32.1  & 25.1  & 9.3   & 9.3   & 15.5  & 10.6  & 19.2  \\
Index Offset            & 5.6   & 5.2   & 4.9   & 2.1   & 2.0   & 2.8   & 2.3   & 3.6   \\
Generate Keys           & 8.5   & 8.0   & 7.1   & 6.3   & 7.6   & 8.3   & 6.7   & 7.5   \\
Sorting                 & 14.6  & 17.5  & 10.6  & 10.0  & 13.4  & 19.0  & 10.9  & 13.7  \\
Tile Ranges             & 0.5   & 0.6   & 0.4   & 0.4   & 0.5   & 0.7   & 0.4   & 0.5   \\
Rasterization Forward   & 31.9  & 28.4  & 20.4  & 17.4  & 23.1  & 34.0  & 22.1  & 25.3  \\
Copy Image to Device    & 12.2  & 13.0  & 12.3  & 20.1  & 19.0  & 19.3  & 19.1  & 16.4  \\
Loss Gradient           & 11.5  & 12.4  & 11.6  & 18.5  & 18.5  & 18.6  & 18.5  & 15.7  \\
Rasterization Backward  & 163.0 & 151.0 & 110.7 & 89.7  & 116.0 & 172.8 & 106.5 & 130.0 \\
Proj Bwd + Optimizer    & 236.2 & 217.5 & 199.1 & 58.6  & 53.2  & 86.3  & 66.7  & 131.1 \\
Densification           & 8.8   & 9.1   & 8.0   & 2.4   & 2.4   & 3.9   & 2.9   & 5.4   \\
Unaccounted             & 49.5  & 50.2  & 47.8  & 38.3  & 37.2  & 42.3  & 38.9  & 43.5  \\
\hline
Total         & 574.9 & 545.1 & 458.1 & 273.0 & 302.2 & 423.7 & 305.6 & 411.8 \\
\hline
\end{tabular}

Timing breakdown on NVIDIA RTX 3090, MCMC 1M densification:

\begin{tabular}{|l|c|c|c|c|c|c|c|c|}
\hline
 & bicycle & garden & stump & bonsai & counter & kitchen & room & Average \\
\hline
Projection Forward      & 7.2   & 8.0   & 7.4   & 8.7   & 8.9   & 9.3   & 8.0   & 8.2   \\
Index Offset            & 1.3   & 1.3   & 1.3   & 1.3   & 1.3   & 1.3   & 1.3   & 1.3   \\
Generate Keys           & 7.7   & 7.7   & 6.9   & 17.4  & 15.0  & 16.0  & 15.4  & 12.3  \\
Sorting                 & 8.4   & 9.3   & 8.6   & 15.1  & 19.4  & 16.9  & 14.1  & 13.1  \\
Tile Ranges             & 0.3   & 0.3   & 0.3   & 0.6   & 0.7   & 0.6   & 0.5   & 0.5   \\
Rasterization Forward   & 18.3  & 14.4  & 16.7  & 29.1  & 37.3  & 30.1  & 29.7  & 25.1  \\
Copy Image to Device    & 12.1  & 12.9  & 12.2  & 20.1  & 19.1  & 19.2  & 18.9  & 16.3  \\
Loss Gradient           & 11.5  & 12.3  & 11.6  & 18.6  & 18.7  & 18.7  & 18.6  & 15.7  \\
Rasterization Backward  & 81.7  & 76.5  & 81.9  & 142.4 & 175.1 & 148.0 & 136.8 & 120.3 \\
Proj Bwd + Optimizer    & 44.0  & 47.0  & 42.4  & 48.1  & 47.5  & 48.7  & 46.3  & 46.3  \\
Densification           & 3.1   & 3.3   & 3.0   & 3.4   & 3.3   & 3.4   & 3.3   & 3.2   \\
Unaccounted             & 22.6  & 22.5  & 22.3  & 23.3  & 23.7  & 23.7  & 23.2  & 23.0  \\
\hline
Total         & 218.2 & 215.5 & 214.6 & 327.9 & 370.0 & 335.7 & 316.1 & 285.4 \\
\hline
\end{tabular}

\newpage

Timing breakdown on AMD Radeon RX 7800 XT, Default densification:

\begin{tabular}{|l|c|c|c|c|c|c|c|c|}
\hline
 & bicycle & garden & stump & bonsai & counter & kitchen & room & Average \\
\hline
Projection Forward      & 71.4    & 67.2   & 60.7   & 24.2   & 23.6    & 33.5    & 26.3  & 43.9  \\
Index Offset            & 7.9     & 12.0   & 6.9    & 9.3    & 9.1     & 9.8     & 9.3   & 9.2   \\
Generate Keys           & 11.7    & 11.3   & 9.8    & 6.4    & 7.6     & 8.6     & 7.0   & 8.9   \\
Sorting                 & 13.8    & 16.5   & 9.9    & 9.4    & 13.0    & 18.2    & 10.4  & 13.0  \\
Tile Ranges             & 0.6     & 0.7    & 0.4    & 0.4    & 0.6     & 0.9     & 0.4   & 0.6   \\
Rasterization Forward   & 67.1    & 69.0   & 42.9   & 38.9   & 51.1    & 79.9    & 46.4  & 56.5  \\
Copy Image to Device    & 282.8   & 353.5  & 384.0  & 618.5  & 557.2   & 611.8   & 553.1 & 480.1 \\
Loss Gradient           & 20.4    & 23.6   & 22.2   & 39.3   & 38.6    & 39.9    & 38.2  & 31.8  \\
Rasterization Backward  & 183.4   & 184.0  & 122.0  & 106.9  & 140.6   & 218.0   & 126.1 & 154.4 \\
Proj Bwd + Optimizer    & 511.1   & 484.7  & 439.6  & 129.1  & 118.3   & 193.1   & 148.8 & 289.2 \\
Densification           & 13.3    & 14.1   & 12.0   & 4.5    & 4.2     & 7.0     & 4.5   & 8.5   \\
Unaccounted             & 18.4    & 18.9   & 17.9   & 16.9   & 16.9    & 17.8    & 16.9  & 17.7  \\
\hline
Total         & 1201.9  & 1255.4 & 1128.4 & 1003.9 & 980.6   & 1238.6  & 987.5 & 1113.8 \\
\hline
\end{tabular}

Timing breakdown on AMD Radeon RX 7800 XT, MCMC 1M densification:

\begin{tabular}{|l|c|c|c|c|c|c|c|c|}
\hline
 & bicycle & garden & stump & bonsai & counter & kitchen & room & Average \\
\hline
Projection Forward      & 15.3  & 16.4  & 14.8  & 17.3   & 17.3   & 17.8  & 16.0  & 16.4  \\
Index Offset            & 7.5   & 7.5   & 7.5   & 7.6    & 7.5    & 7.5   & 7.5   & 7.5   \\
Generate Keys           & 6.3   & 6.2   & 6.3   & 13.9   & 12.5   & 12.9  & 13.0  & 10.2  \\
Sorting                 & 7.7   & 8.3   & 7.8   & 14.5   & 18.8   & 16.5  & 13.3  & 12.4  \\
Tile Ranges             & 0.3   & 0.4   & 0.3   & 0.6    & 0.9    & 0.8   & 0.6   & 0.6   \\
Rasterization Forward   & 37.6  & 34.4  & 34.6  & 62.1   & 80.2   & 68.8  & 60.2  & 54.0  \\
Copy Image to Device    & 340.1 & 321.3 & 248.7 & 688.9  & 556.8  & 485.4 & 524.4 & 452.2 \\
Loss Gradient           & 21.1  & 22.3  & 19.7  & 40.9   & 38.9   & 38.1  & 38.1  & 31.3  \\
Rasterization Backward  & 100.3 & 90.1  & 93.6  & 169.8  & 215.5  & 184.4 & 161.0 & 145.0 \\
Proj Bwd + Optimizer    & 95.0  & 100.5 & 91.5  & 103.6  & 101.5  & 103.6 & 99.6  & 99.3  \\
Densification           & 9.9   & 10.5  & 9.4   & 10.7   & 10.6   & 10.8  & 10.5  & 10.4  \\
Unaccounted             & 13.4  & 13.7  & 13.5  & 14.2   & 13.9   & 14.1  & 14.0  & 13.8  \\
\hline
Total         & 654.6 & 631.5 & 547.6 & 1144.4 & 1074.4 & 960.8 & 958.3 & 853.1 \\
\hline
\end{tabular}

The largest time difference betwen AMD Radeon RX 7800 XT and NVIDIA RTX 3090 is copying image to device, which is nearly 30 times slower on AMD on average. We believe this is due to a PCIe difference, and the end-to-end training time can be optimized with asynchronous data transfer.